# Learning across label confidence distributions using Filtered Transfer Learning

Seyed Ali Madani Tonekaboni[1,2,3,4], Andrew E. Brereton[1], Zhaleh Safikhani[1], Andreas Windemuth[1], Benjamin Haibe-Kains[2,3,4,5], Stephen MacKinnon[1]

[1]Cyclica, Toronto, ON, Canada
[2]Department of Medical Biophysics, University of Toronto, Toronto, ON, Canada
[3]Princess Margaret Cancer Centre, Toronto, ON, Canada
[4]University Health Network, Toronto, ON, Canada
[5]Vector Institute, Toronto, ON, Canada

## Abstract

Performance of neural network models relies on the availability of large datasets with minimal levels of uncertainty. Transfer Learning (TL) models have been proposed to resolve the issue of small dataset size by letting the model train on a bigger, task-related reference dataset and then fine-tune on a smaller, task-specific dataset. In this work, we apply a transfer learning approach to improve predictive power in noisy data systems with large variable confidence datasets. We propose a deep neural network method called *Filtered Transfer Learning* (FTL) that defines multiple tiers of data confidence as separate tasks in a transfer learning setting. The deep neural network is fine-tuned in a hierarchical process by iteratively removing (filtering) data points with lower label confidence, and retraining. In this report we use FTL for predicting the interaction of drugs and proteins. We demonstrate that using FTL to learn stepwise, across the label confidence distribution, results in higher performance compared to deep neural network models trained on a single confidence range. We anticipate that this approach will enable the machine learning community to benefit from large datasets with uncertain labels in fields such as biology and medicine.

# 1 Introduction

The availability of large domain-specific datasets has made it possible for the corresponding industries and scientific communities to benefit from deep neural network modeling in target tasks. The complexity of neural network models necessitates collection of massive numbers of data points having both feature and output values for supervised learning tasks. Increasing dataset size can result in improvements in performance of machine learning (including neural network) models (Crichton et al. 2017; Jaehoon Lee, Yasaman Bahri, Roman Novak, Samuel S. Schoenholz, Jeffrey Pennington, Jascha Sohl-Dickstein 2018). Although there could be a limit to performance improvement gained by increasing dataset size, consistently improving performance is still observed from increases in dataset size when using deep neural network models (Jaehoon Lee, Yasaman Bahri, Roman Novak, Samuel S. Schoenholz, Jeffrey Pennington, Jascha Sohl-Dickstein 2018). Although large datasets for supervised learning tasks have become available in many domains, not all data points have the same level of confidence regarding the truth of their assigned labels.

Although output values (either categorical, for classification, or continuous, for regression) are available for each data point in most datasets, the confidence level for output values need not be the same. In aggregated data environments, confidence in the individual data points may vary in a quantifiable manner by primary data source or measurement type. For example, consider a critical task in the pharmaceutical industry and biomedical research community: developing machine learning models for predicting the interaction of proteins and drugs. These models rely on datasets like STITCH (Szklarczyk et al. 2016), which aggregate interactions from multiple primary sources and report confidence values based on a composite approach to defining certainty (Szklarczyk et al. 2016). Differences in label confidence make model building challenging, as the optimization cannot be done while amalgamating all the data points in the training process. The challenge of mixed confidence training data is not restricted to the domain of protein and drug interaction; in practice, data labeling is done based on either computational algorithms or human experts (or even non-experts), and neither approach is perfect. Other examples with differences in data point label confidence include: radiological or histopathological images or image segment labels (Brady 2017; Oakden-Rayner 2020), and measured resistance to cancer drugs (Hafner et al. 2016).

Extracting benefit from data points that have variable confidence regarding their true labels is a challenging task at best. This problem has been tackled previously using confidence based weight-assignment in the optimization process (Reamaroon et al. 2019; Almeida et al. 2020; Hagenah, Leymann, and Ernst 2019; Luo, Dang, and Chen 2017). However, some systems may be constrained from using weight-based strategies, such as: those with discrete confidence tiers rather than continuous probabilities, systems with only partial confidence assignments, or systems using simulated random negatives. A good example of such cases is the interaction of drugs and proteins. There are large datasets like STITCH (Szklarczyk et al. 2016) containing millions of data points that can be treated as positive labels (true interactions). While true negative data points are not as readily available, they can be randomly generated as a subset of all possible drug-protein interactions that are not reported as positives in STITCH. In this case, the negatives do not have any associated confidence, demonstrating a need for an alternative approach for handling variable confidence data points during model training.

Here we propose a new modeling technique called Filtered Transfer Learning (FTL) to benefit from variable confidence data points. FTL is a step-wise transfer learning method for learning across confidence distributions, building on the already established concept of transfer learning (Yosinski et al. 2014). To illustrate key principles of the approach, the Filtered Transfer Learning was applied to a sample system for predicting drug-target interactions.

## 2 Datasets

The source dataset of drug-target interactions was retrieved from STITCH v5.0 using the web-based interface, selecting human-only interactions and the complete STITCH v5.0 compounds dataset, with 15,473,939 and 116,224,359 records respectively (Szklarczyk et al. 2016). Only compounds ("drugs") with known human interactions (786,494) were parsed and featurized using RDKIT morgan2 circular fingerprints with 1024 bits [version 2018.9.1] (Landrum et al. 2020); 52 compounds were excluded due to errors in the featurization step. A snapshot of reviewed protein records ("targets") corresponding to the human canonical proteome (up000005640) was retrieved from uniprot on February 18th, 2020 ("UniProt: A Worldwide Hub of Protein Knowledge" 2019). The STITCH v5.0 interactions were mapped to 13,900 corresponding uniprot proteins via their String IDs. Among the mapped proteins were 5508 distinct Pfam domains (El-Gebali et al. 2019). Simple feature bit vectors were created for each protein by vectorizing the presence/absence of each Pfam domain in accordance with a fixed-length index, a strategy previously-described by Secure-DTI (Hie, Cho, and Berger 2018). The resulting datasets had 786,442 distinct compound bit vectors (1024-bits), 13,900 distinct protein bit vectors (5508-bits), and 12,152,512 filtered pairwise interactions with variable (labeled) confidence scores provided by STITCH v5.0. Features for positive training examples were obtained by concatenating the corresponding compound and protein bit vectors of the filtered pairwise interactions, while negative training examples were obtained by randomly selecting compound-protein pairs and concatenating their respective bit vectors if the pair is not already a positive interaction. Further details, including the data processing scripts and all derived bit vectors are available in the associated GitHub repository.

## 3 Methods

### 3.1 Dimensionality reduction

The binary features (bit vectors) of protein and chemical space are reduced to 128 and 64 latent variables using variational autoencoders. The variational autoencoders are fully connected neural networks with [2048, 512, 128] and [256, 128, 64] encoder architectures for protein and chemical space, respectively (Table 1). Both were optimized using the Adam optimizer for 500 epochs (convergence) (Table 1).

**Table 1**. Details of variational autoencoder models used to define separate latent spaces for protein and chemical space.

| Model hyperparameters | Protein space | Chemical space |
|---|---|---|
| Total number of data points | 13,900 | 786,442 |
| Number of input features | 5508 | 1024 |
| Encoder layers (fully connected) | [2048, 512, 128] | [256, 128, 64] |
| Number of epochs | 500 | 500 |
| Batch size | 1000 | 1000 |
| Optimization algorithm | Adam | Adam |
| Learning rate | 0.0001 | 0.0001 |

### 3.2 Neural network model for drug-protein interaction

The latent space representations of drugs and proteins were used as the input feature space in a fully connected feed-forward neural network architecture: 192 features (128 and 64 latent variables of proteins and drugs, respectively), are used in an architecture of [128, 64, 32, 16, 8] fully connected layers for binary prediction of interaction between drugs and proteins. The Adam optimization algorithm (Kingma and Ba 2014) with a learning rate of 0.001 was used to optimize the neural network model.

### 3.3 Filtered Transfer Learning

Filtered Transfer Learning (FTL) is defined as a neural network trained across confidence distributions in a stepwise process (Figure 1). The neural network is first trained in the lower confidence part of the dataset, which includes the majority of the data points, and then continues to be trained on the higher confidence part of the dataset after filtering out the lower confidence data points (Figure 1). The preliminary training on the low confidence data points can also be done in multiple steps ((n+1)-step FTL; n: number of steps in low confidence data).

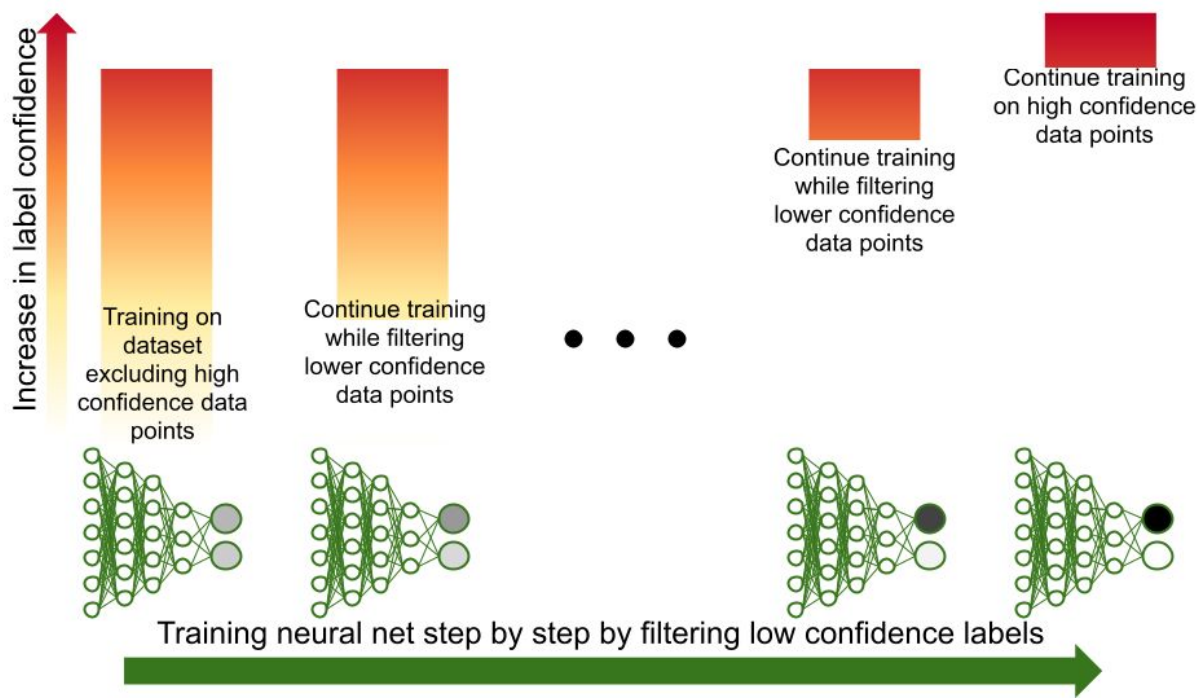

**Figure 1**. Schematic representation of learning stepwise across label confidence distributions by Filtered Transfer Learning (FTL).

## 4 Results

To study the performance of our proposed technique, Filtered Transfer Learning (FTL), we used prediction of drugs and proteins as the target task. Initially, the dimensionality of protein and chemical spaces were reduced using variational autoencoders (Table 1), which reduced computational cost and complexity for the supervised modeling using FTL. The identified latent variables were used for the supervised modeling throughout this study.

The STITCH dataset reports a confidence score for each record of a drug/protein interaction. We used the highest confidence range of this score [900, 1000) as the validation set throughout the manuscript and split the rest to assess performance of supervised models. The confidence range of [700, 900) has been also reported by STITCH as the high confidence range (Szklarczyk et al. 2016). Based on this, validation performance was assessed for a neural network model with [128, 64, 32, 16, 8] layer sizes trained on [700, 900), as well as for models trained on lower confidence ranges, chosen based on percentiles in the confidence distribution (Table 2; Figure 3A). The models reached convergence after 200 epochs (Figure 3B). Their loss (Figure 3B) and accuracy (Figure 3C) on the validation set were compared. The best validation loss and accuracy of the model trained on the high confidence range [700, 900) performed better than the models trained on lower confidence ranges (Table 3).

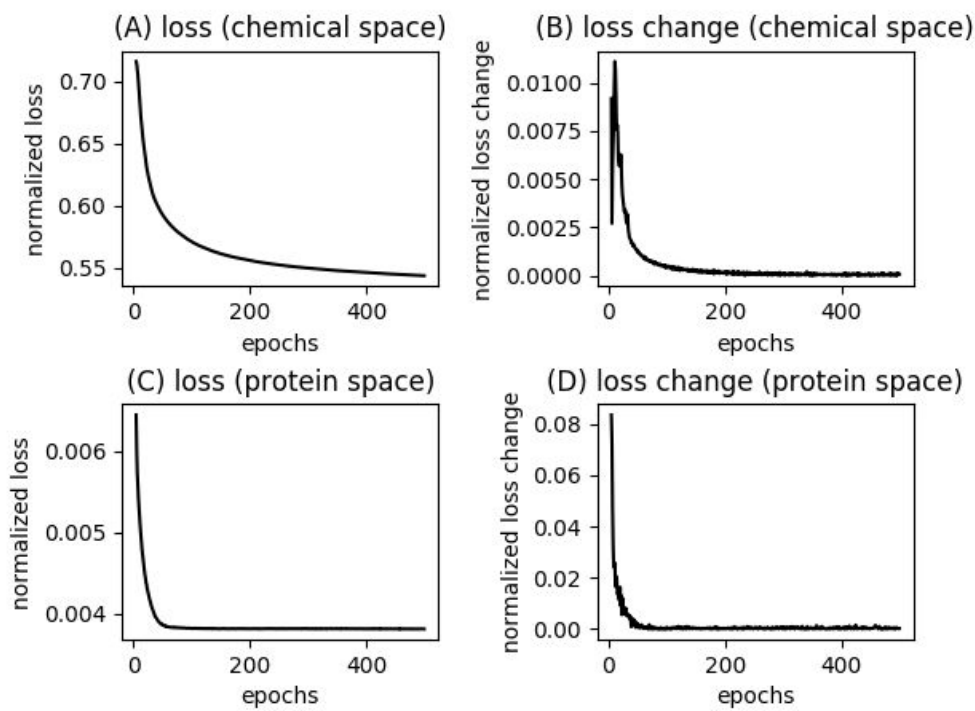

**Figure 2**. Loss (A, C) and Loss Change (B, D) per epoch for optimization of the designed variational autoencoder to reduce dimensionality of protein (C, D) and chemical (A, B) space.

**Table 2**. STITCH score cutoffs considered for training the neural network models used for predicting high confidence interactions (score: [900, 1000)) between drug/protein pairs in the STITCH database.

| Cutoff range for confidence of positive interactions in STITCH database | Corresponding percentile range | Number of positive data points (interactions) |
|---|---|---|
| [319,900) | [82,100) | 2,186,942 |
| [389,900) | [90,100) | 1,204,936 |
| [700,900) | [98,100) | 249,278 |

To assess whether FTL can improve the performance of the neural network modelling, 2-step FTLs were developed using the [700,900) positive label confidence range for the second training step. Using the [319, 700) positive label confidence range as the first step of the FTL model training resulted in lower loss (5.203e-5) and higher accuracy (79.759) compared to the neural network model trained only on the [700, 900) confidence range (Figure 4; Table 3 and 4). The confidence range of [319, 700) was then split into two equal parts by number of data points, and two other 2-step FTLs were trained using the resulting ranges of [319, 389) and [389, 700). Both new FTLs outperformed the model trained only on the [700, 900) confidence range (Figure 4; Table 3 and 4). However, the FTL with [389, 700) as the first step had lower loss and higher accuracy (loss=5.317e-5; accuracy =79.203; Table 4) compared to the FTL with the lower confidence range of [319, 389) for training the first step (loss=5.424e-5; accuracy =78.549; Table 4)**.**

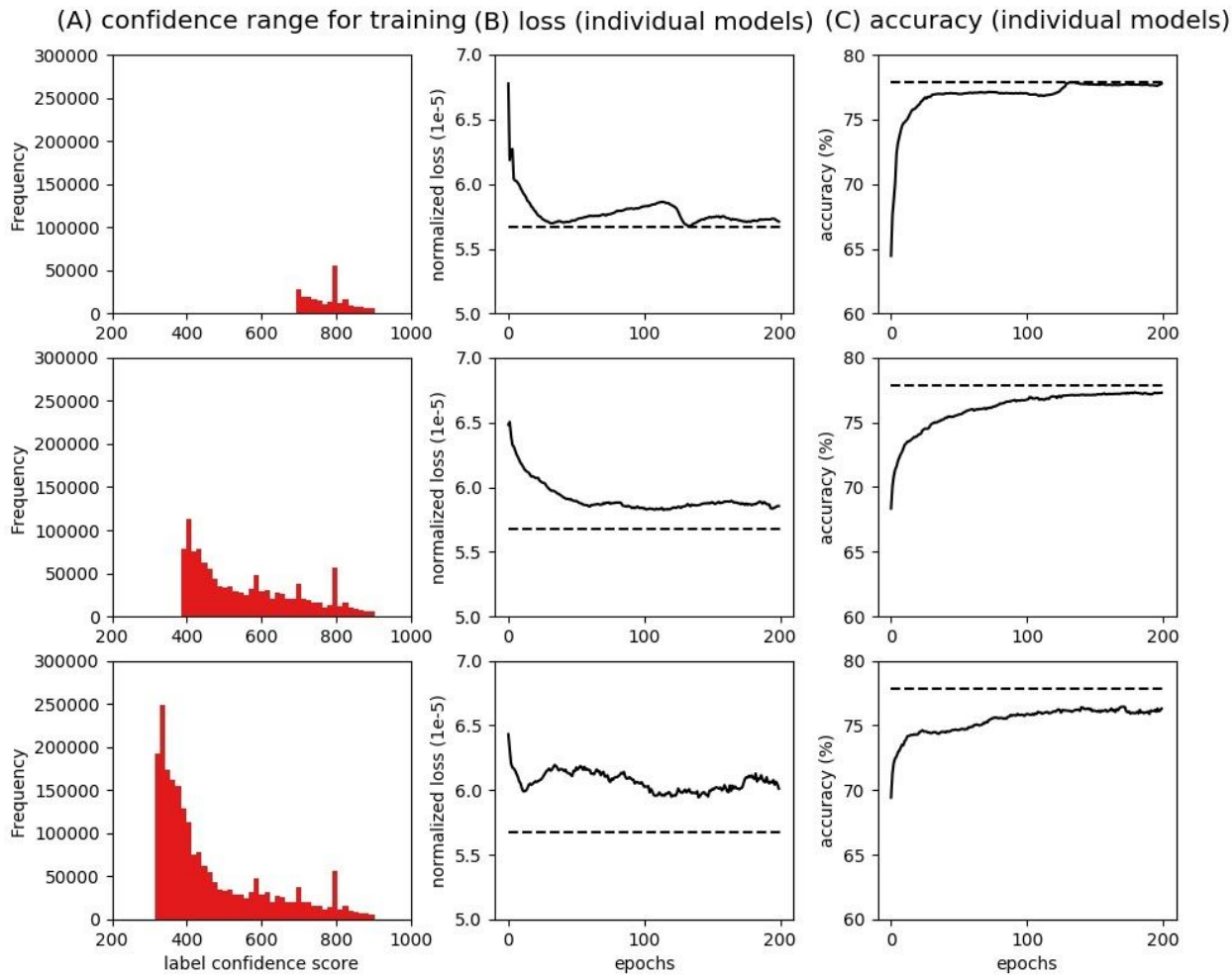

**Figure 3**. (A) Positive label confidence range used in training a neural network model to predict the positive interactions with the label confidence score range of [900, 1000). (B) Validation loss and (C) accuracy of the neural network models trained on their corresponding label confidence range shown in panel (A).

**Table 3.** Validation loss and accuracy of the neural network model trained on specified positive label confidence range in each row and validated on [900, 1000) positive label confidence range.

| Training confidence range | Validation loss | Validation accuracy (%) |
|---|---|---|
| [319,900) | 5.943e-5 | 76.444 |
| [389,900) | 5.824e-5 | 77.321 |
| [700,900) | 5.673e-5 | 77.874 |

In the transition from step one to step two in the training of two-step FTLs, only the neural network weights were left as available degrees of freedom. To better understand the difference between the two steps of the FTLs, we identified the Euclidean distance (normalized to the number of weights) between weights of each layer at the end of 100 epochs of training in step one of the FTL and after initial 20 epochs of training in step 2 (Figure 5). We then compared these distances with the distances of weights of layers after 100 and 120 epochs of training using the same confidence range used for step 1 of the 2-step FTL (Figure 5). This comparison revealed that the network continues changing the weights of the first 2 layers in its training going from step 1 to step 2 while it drastically slows down the rate of change in the weights of last 3 layers in this transition (Fold change < 0.55; Figure 5).

**Table 4.** Validation loss and accuracy of the 2-step FTLs trained on specified positive label confidence ranges and validated on the [900, 1000) positive label confidence range.

| Training confidence range | Validation loss | Validation accuracy (%) |
|---|---|---|
| [319,700)->[700, 900) | 5.203e-5 | 79.759 |
| [389,700)->[700, 900) | 5.317e-5 | 79.203 |
| [319,389)->[700, 900) | 5.424e-5 | 78.549 |

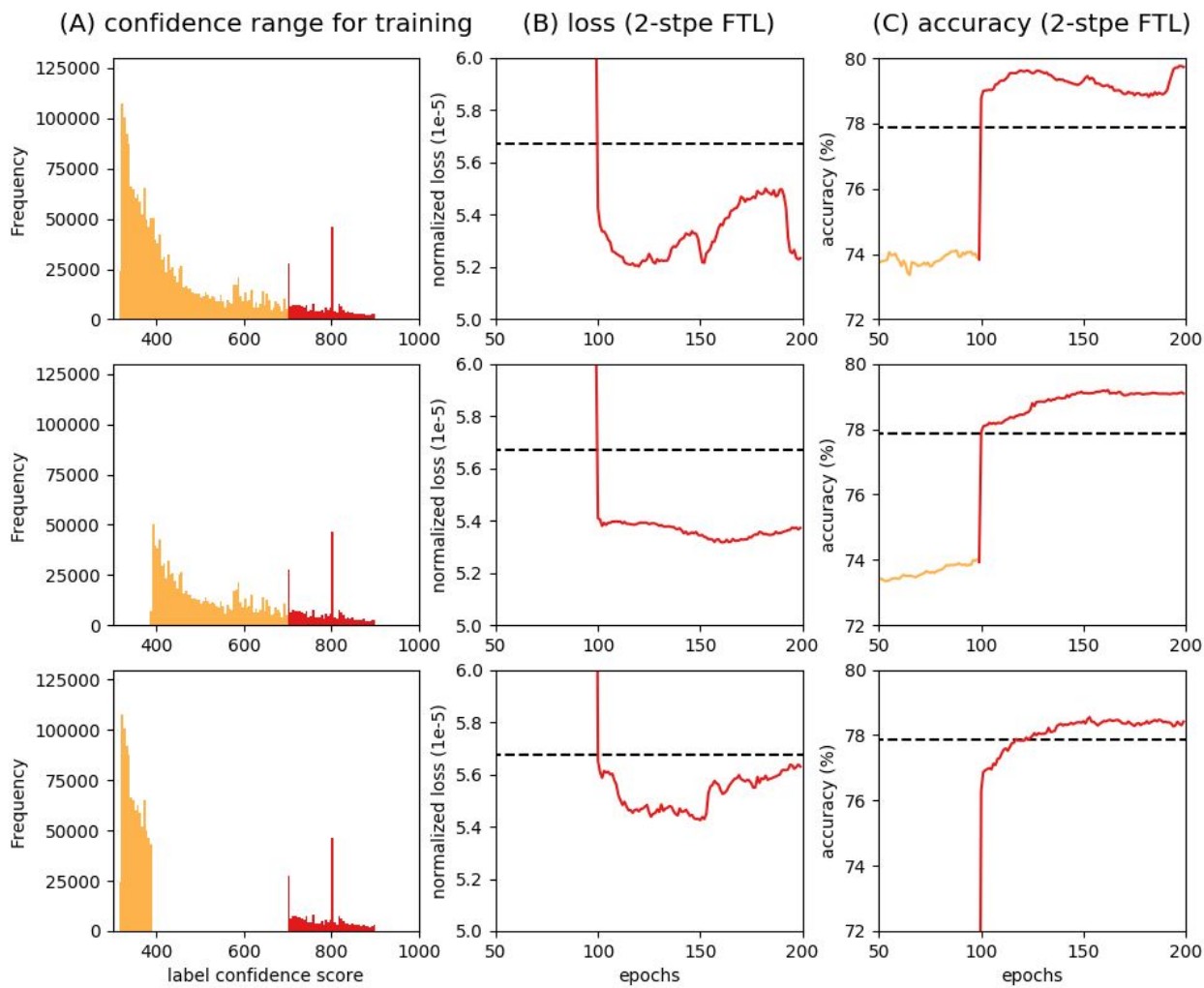

**Figure 4**. (A) Positive label confidence range used in training a neural network model for predicting the positive interactions with the label confidence score range of [900, 1000). Validation (B) loss and (C) accuracy of the FTL trained first on label confidence ranges of either [319,700), [389, 700) or [319, 389) for 100 epochs, as shown in (A), then trained for another 100 epochs on the positive interactions with the label confidence range of [700, 900). The horizontal dashed lines depict the best validation loss (B) or accuracy (C) of neural network models trained only on one label confidence range [700, 900) (Table 3).

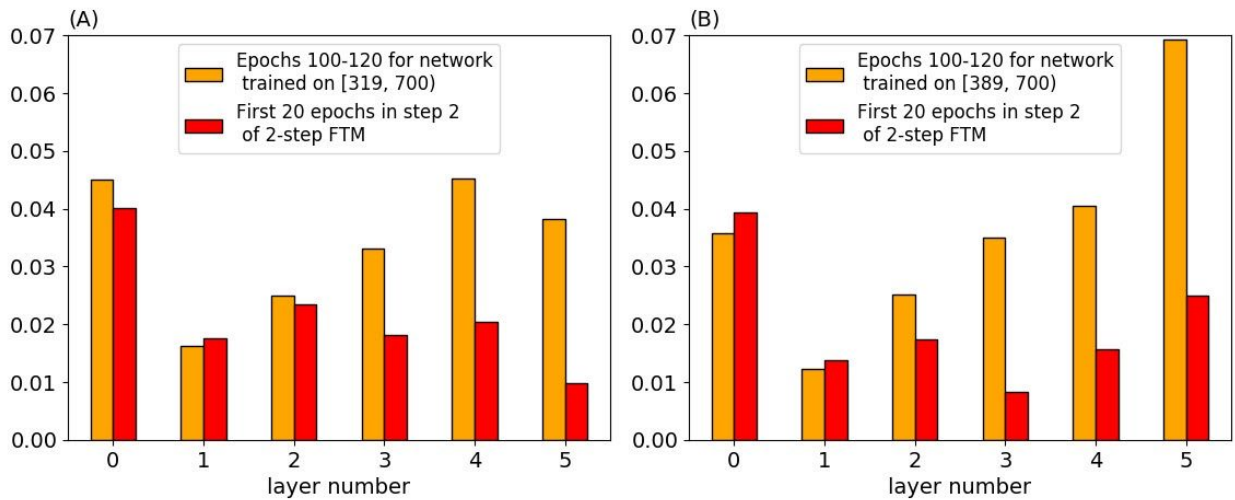

**Figure 5**. Distance between the weights of layers of neural network models either between epochs 100-120 of the network trained on one confidence range or between epoch 100 of step 1 and epoch 20 of step 2 in 2-step FTL. (A) Confidence range: [319, 700) (orange); 2-step FTL: [319,700)->[700,900) (red); (B) Confidence range: [389, 700) (orange); 2-step FTL: [389,700)->[700,900) (red).

**Limitations of this study**. Although the proposed model in this study (Filtered Transfer Learning) provides promising results for predicting drug-target interactions, we acknowledge that the study has some limitations:

1) Since the developed variational autoencoder models for protein and chemical space are not optimized across hyperparameter space, overall performance likely could still be improved. However, as the same latent variables are used for the training of each neural network model, the identified latent variables helped us to lower computational cost of the study without uniquely favoring FTL or non-FTL models.
2) The results associated with different label confidence range cutoffs cannot be applied directly to other datasets. For example, in the case of image label confidence, the target dataset should be explored to determine appropriate cutoffs to build the training sets.
3) The optimal number of epochs for each label confidence range in FTL should be identified (eg. through a grid search) which was not explored in this study.
4) There is no correction for compound-series bias in the validation test set( i.e. the performance measures may be an overestimate because of very similar compounds in training and validation sets).

Nonetheless, while the specific models presented in this study are not optimized for practical applications, we believe this sample system clearly demonstrates the application of a Filtered Transfer Learning towards improving predictive performance on variable confidence data systems.

## 5 Conclusion

Noisy datasets are commonplace among practical systems that apply deep learning strategies. Optimizing performance of real-world predictions requires the modelling system to maximize the value of all available data. While noisier data may contain incorrect entries, the presence of a predictive signal indicates a net-positive informational value. Strategies for dealing with variable-confidence training data present an alternative to data sanitization techniques that exclude noisier data in bulk, and thus discard useful information. However, many such strategies, like ensemble learning or training weights, may impose functional disadvantages or technical constraints, limiting their overall practicality.

We proposed a new approach called Filtered Transfer Learning (FTL) to benefit from data points with variable label confidence while training a neural network. FTL is trained across label confidence distributions in a stepwise manner starting from low confidence and moving to high confidence data points (with or without overlapping data points across the steps). We demonstrate the utility of our approach using a dataset for predicting interactions between drugs and proteins, a matching problem with sparsely populated positive examples, a lack of explicit negative examples, and defined label confidence. The presented results show higher performance for FTL when compared to neural network models trained on the data points with different label confidence levels in a single step. To study the effect of label confidence on the trained neural networks, we compared the weights of neural networks between the steps of FTL models. The weights of the first layers including the weights connecting the input layer to the first hidden layer kept changing while changes in the last layers slowed down dramatically during transition between the steps in FTL models. By presenting a new deep learning strategy for dealing with variable confidence training data, we challenge the data *quality vs quantity* tradeoff. The Filtered Transfer Learning has minimal additional technical constraints and leverages the breadth of larger datasets, without diluting the impact of richer training examples.

### Acknowledgment

We express our sincere gratitude to Mitacs and their Vector-Institute collaboration program for supporting this research through their internship program. We also thank the Cyclica Research and Development team for critical feedback throughout the development progress.